\documentclass[preprint,12pt]{elsarticle}

\usepackage{amssymb}
\usepackage{amsmath}

\journal{Biomedical Signal Processing and Control}

\begin{document}

\begin{frontmatter}



\title{GCA-ResUNet:Image segmentation in medical images using grouped coordinate attention}


\author{Ding Jun} 
\author[1,*]{Gao Shang}

\affiliation{organization={Jiangsu University of Science and Technology},
            addressline={No. 666 Changhui Road, Zhenjiang, Jiangsu, P.R. China}, 
            city={zhenjiiang},
            postcode={212003}, 
            state={jiangsu},
            country={China}}

\cortext[*]{Corresponding author: Gao Shang, Email: gao\_shang@just.edu.cn}

\begin{abstract}
Medical image segmentation underpins computer-aided diagnosis and therapy by supporting clinical diagnosis, preoperative planning, and disease monitoring. While U-Net-style CNNs excel via encoder–decoder designs with skip connections, they struggle to capture long-range dependencies; Transformer variants address global context but incur heavy computation and data demands. This paper proposes GCA-ResUNet, a efficient segmentation network that integrates  Grouped Coordinate Attention (GCA) into ResNet-50 residual blocks. GCA employs grouped coordinate modeling to jointly encode global dependencies across channels and spatial locations, strengthening feature representation and boundary delineation with minimal parameter and FLOP overhead compared with self-attention. On Synapse, GCA-ResUNet achieves a Dice score of 86.11\%; on ACDC, it reaches 92.64\%, surpassing several state-of-the-art baselines while maintaining fast inference and favorable computational efficiency. These results indicate that GCA provides a practical route to endow convolutional architectures with global modeling capability, yielding high-accuracy, resource-efficient medical image segmentation.
\end{abstract}


\begin{highlights}
\item A lightweight hybrid segmentation network, GCA-ResUNet, that couples ResNet-based local feature extraction with explicit global dependency modeling, preserving encoder–decoder efficiency while enhancing long-range context and boundary fidelity.
\item A plug-and-play Grouped Coordinate Attention (GCA) module that combines channel grouping with direction-aware global pooling to realize low-complexity global modeling, strengthening cross-channel semantics and fine-grained spatial cues with minimal parameter/FLOP overhead.
\item An effective multi-stage integration strategy that inserts GCA into encoder and multi-scale fusion paths, yielding improved delineation of complex structures and blurred boundaries under limited computational budgets.
\item Comprehensive experiments on multiple medical segmentation benchmarks demonstrating consistent gains in accuracy and robustness.
\end{highlights}


\end{frontmatter}



\section{Introduction}
Medical image segmentation underpins computer-aided diagnosis and therapy by providing pixel-level delineation of organs and lesions for diagnosis, treatment planning, and longitudinal monitoring. Despite substantial progress, 
many clinical scenarios still feature low contrast, heterogeneous textures, and ambiguous boundaries that challenge robust delineation. Practical deployment additionally requires models to balance accuracy with computational and memory efficiency, especially on resource-constrained platforms. These constraints motivate architectures that retain global context while remaining lightweight and stable.

Early mainstream approaches are CNN-based, exemplified by U-Net and its variants\cite{ronneberger2015unet,gu2019cenet,li2023eresunetpp}, which fuse multi-scale features through a symmetric encoder–decoder with skip connections. While effective for local pattern extraction, standard convolutions operate with limited receptive fields and implicit long-range interactions, leading to missed global dependencies. As a result, segmentations may suffer from boundary blurring, topology breakage, and sensitivity to low-contrast regions, particularly for small or thin structures.

Transformer-based routes\cite{vaswani2017attention} were introduced to remedy global modeling limitations, either by replacing convolutional stages or forming hybrids (e.g., TransUNet\cite{chen2024transunet} , Swin-UNet\cite{cao2022swinunet} , and subsequent designs\cite{zhu2024fullunet,islam2024costunet}). These models explicitly model long-range dependencies and cross-scale interactions, often improving accuracy on complex anatomy. However, they typically incur high parameter counts and FLOPs, stronger data requirements, and longer training and inference times, which hinder use in settings with limited compute or memory. Moreover, gains can diminish when datasets are small or domain shifts are pronounced.

A third line exploits lightweight attention within CNNs to inject global cues at low cost. SE\cite{hu2018senet} adaptively reweights channels, CBAM\cite{woo2018cbam} couples channel and spatial attention, and Coordinate Attention\cite{hou2021coordatt} encodes positional information into channel attention to enhance direction-aware context. Although these modules improve discriminability with modest overhead, they may still insufficiently model global dependencies across channels and space, or lack explicit mechanisms to preserve fine-grained boundary details in challenging cases.

We propose GCA-ResUNet, a hybrid network that integrates Grouped Coordinate Attention (GCA) into a ResNet-UNet\cite{alabdulhafith2024resnetunet} backbone. GCA employs channel-wise grouping to capture cross-channel global dependencies with low complexity and applies direction-aware global pooling along horizontal and vertical axes to encode structural correlations in space. Inserted at encoder, decoder, and multi-scale fusion stages, GCA strengthens long-range context, cross-channel interaction, and boundary recognition while incurring minor parameter and FLOP overhead compared with self-attention. Experiments on multiple benchmarks show that GCA-ResUNet attains competitive accuracy and efficiency, particularly for complex anatomy and blurred contours, indicating a practical path to endow convolutional architectures with effective global modeling for medical image segmentation.

In summary, the main contributions of this paper are as follows:
\begin{itemize}
    \item A lightweight hybrid segmentation network, GCA-ResUNet, that couples ResNet-based local feature extraction with explicit global dependency modeling, preserving encoder–decoder efficiency while enhancing long-range context and boundary fidelity.
    \item A plug-and-play Grouped Coordinate Attention (GCA) module that combines channel grouping with direction-aware global pooling to realize low-complexity global modeling, strengthening cross-channel semantics and fine-grained spatial cues with minimal parameter/FLOP overhead.
    \item An effective multi-stage integration strategy that inserts GCA into encoder and multi-scale fusion paths, yielding improved delineation of complex structures and blurred boundaries under limited computational budgets.
    \item Comprehensive experiments on multiple medical segmentation benchmarks demonstrating consistent gains in accuracy and robustness.
\end{itemize}

\section{Related work}

\subsection{CNN-based Medical Image Segmentation}

Convolutional neural networks (CNNs) have long dominated medical image segmentation due to their structure-based bias induction, parameter sharing mechanisms, and efficient modeling of local patterns. Among these, U-Net stands as a representative model. Through its symmetric encoder-decoder architecture and skip connections, it achieves abstraction of high-level semantic information while preserving low-level details, effectively integrating multi-scale features. This enables its outstanding performance across diverse medical segmentation tasks, including organs, tumors, and blood vessels.

Building upon U-Net, researchers have proposed various enhancement strategies to further strengthen feature representation and training stability. For instance, incorporating more powerful backbone networks like ResNet\cite{he2016resnet} and DenseNet\cite{huang2017densenet} improves feature extraction capabilities and mitigates gradient vanishing issues. However, despite CNNs' inherent advantage in modeling local patterns, their receptive fields remain constrained by kernel size and network depth. This limitation impairs their ability to capture long-range dependencies and global contextual information. This issue becomes particularly pronounced when handling complex spatial structures, low-contrast tissues, or regions with blurred boundaries, often resulting in unclear segmentation boundaries, structural discontinuities, or cross-organ misclassifications.

To further enhance CNN-based segmentation performance, some research-
ers have attempted to introduce more efficient feature fusion and context modeling mechanisms within pure CNN frameworks. For instance, the network proposed by Chen et al\cite{chen2024lightmlp} employs large-kernel depthwise separable convolutions and a Local Feature Weighted Fusion MLP within the U-Net architecture. This approach significantly reduces parameter count and computational complexity while effectively expanding the receptive field and enhancing global information interaction capabilities. This model achieves a balance between performance and efficiency across multiple medical image segmentation datasets, demonstrating the strong potential of convolutional architectures even without relying on Transformers.

To address CNNs' inherent limitations in global modeling, recent research has further explored integrating non-local mechanisms (Non-local Networks)\cite{wang2018nonlocal}, graph neural networks (GNNs)\cite{scarselli2008gnn,kipf2016gcn}, and Transformers into CNN structures. Through self-attention mechanisms, these approaches model long-range dependencies, enabling complementary integration of local features and global context. Notable examples include Swin-UNet and TransUNet, which retain CNN's efficient local modeling capabilities while enhancing global feature perception via Transformers, achieving significant performance gains in multi-organ and complex-structure medical image segmentation.

Inspired by this, this paper proposes embedding the Grouped Coordinate Attention (GCA) module into traditional CNN architectures. While maintaining low computational overhead, GCA balances modeling local details and global dependencies, thereby effectively enhancing the accuracy and robustness of medical image segmentation.

\subsection{Transformer-based Segmentation Networks}

To overcome the locality constraints of convolutions, the Transformer architecture~\cite{shen2024imagpose,shen2025imagdressing,shen2025imaggarment,shenlong} rapidly gained prominence in visual tasks through its self-attention mechanism. Vision Transformers (ViTs) efficiently model long-range feature dependencies by partitioning input images into patches and computing attention weights between features across the entire image. Subsequently, numerous studies introduced Transformers into medical image segmentation, establishing segmentation frameworks centered on global modeling.

Classic works like TransUNet combine the strengths of convolutional and Transformer architectures: first using CNNs to extract low-level local features, then employing ViT to capture global contextual information, achieving breakthrough performance in multi-organ segmentation tasks. Swin-UNet adopts a hierarchical shifted window self-attention mechanism, balancing local and global feature interactions while maintaining computational control, making it suitable for high-resolution medical image processing. UNETR\cite{hatamizadeh2022unetr} further employs a pure Transformer as the encoder, achieving cross-slice semantic fusion through 3D self-attention, significantly enhancing modeling capabilities in 3D MRI/CT scenarios. Despite their excellence in global feature modeling, these approaches reveal several limitations: (1) The computational complexity of self-attention mechanisms scales quadratically with input resolution, substantially increasing memory and time costs; (2) Transformer models lack the local inductive bias inherent in convolutions, making them prone to overfitting on small medical datasets; (3) Their massive parameter size and unstable training process limit deployment on resource-constrained clinical devices. Therefore, reducing computational burden while preserving global modeling capabilities has become a core challenge in current research.

\subsection{Hybrid Architectures and Our Motivation}

Prior to the emergence of CNN–Transformer hybrid architectures in recent years, a series of attention modules had been proposed to introduce low-cost global modeling capabilities into convolutional networks. For instance, Squeeze-and-Excitation (SE) achieves channel re-calibration through squeeze–excitation operations; the Convolutional Block Attention Module builds upon this by jointly modeling channel and spatial dimensions to highlight discriminative regions; Coordinate Attention (CoordAtt) embeds spatial coordinate information into channel attention to gain directional awareness. These modules demonstrate that introducing moderate global information while maintaining manageable computational overhead can significantly enhance the feature discriminative power of convolutional networks, laying the theoretical and engineering foundation for subsequent hybrid architectures.

Recent hybrid approaches like FC-UNetTR\cite{zhu2024fcunettr} and CoST-UNet\cite{islam2024costunet} further combine convolutional encoder/decoders with Transformer-style global modeling to better capture long-range dependencies and multi-scale semantics. While these methods demonstrate excellent accuracy, the introduced attention or tokenization modules typically incur significant computational and memory overhead.

Inspired by these works and aiming to address their limitations, this paper introduces Grouped Coordinate Attention (GCA), a hybrid module that incorporates grouped global guidance in the channel dimension and integrates coordinate encoding in the spatial dimension. GCA preserves the efficient local modeling capabilities of CNNs while introducing scalable global context modeling at a reduced parameter and computational cost. Based on this approach, we further embed the GCA module into the residual units of ResNet50-UNet to construct the GCA-ResUNet network. By reducing the number of ResNet50 layers, removing the fully connected classification head, and introducing the lightweight GCA module, we decrease computational overhead, making the overall network more efficient and suitable for small-sample and resource-constrained scenarios in medical image segmentation.

\section{Methods}

\subsection{Main Network Structure}

\begin{figure*}[t]
    \centering
    \includegraphics[width=0.95\textwidth]{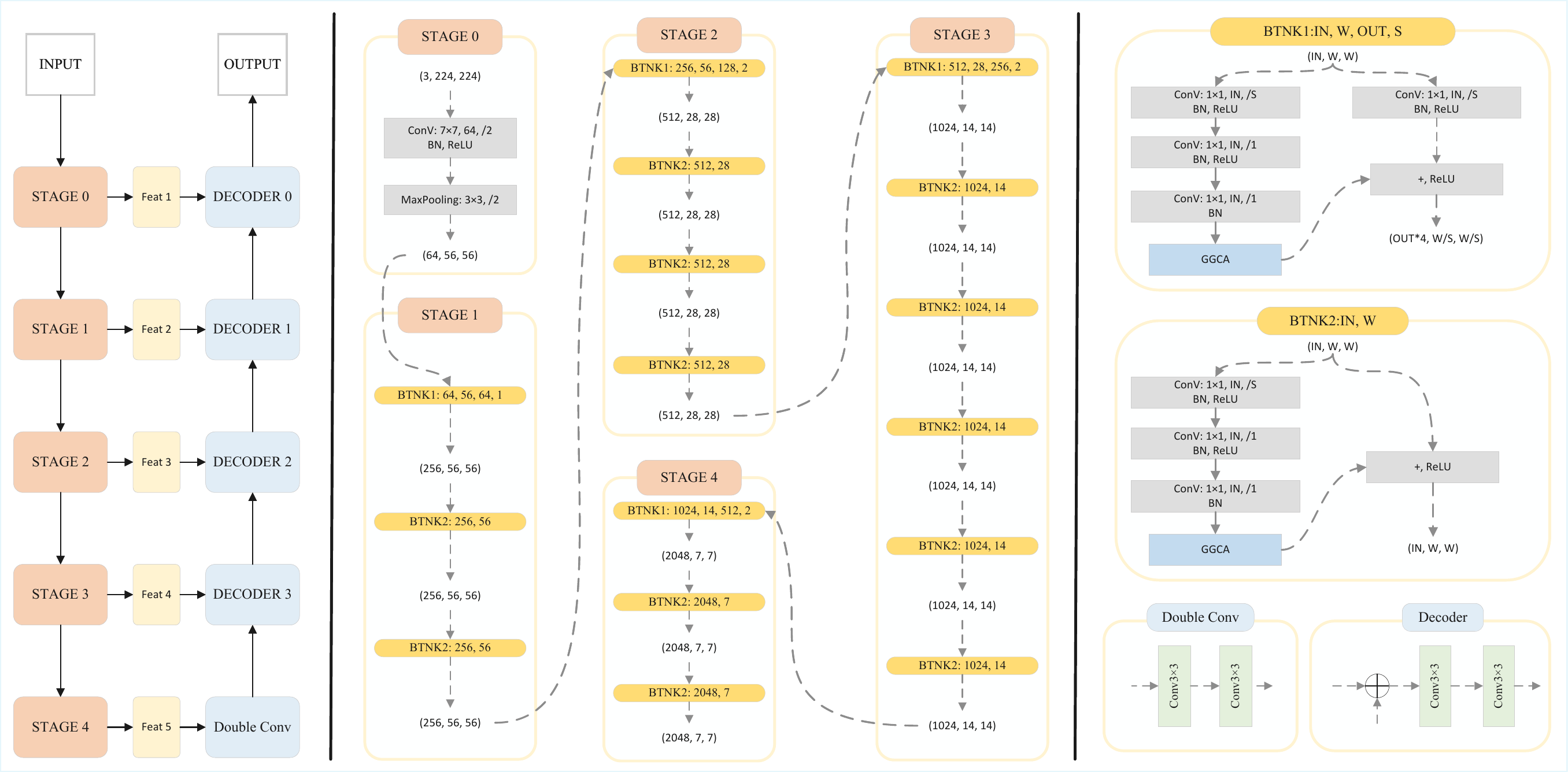}
    \caption{Schematic diagram of the GCA-ResUNet network. The architecture adopts a U-Net–style encoder–decoder structure, with residual blocks incorporated in the encoder and skip connections to enable multi-scale feature extraction and precise pixel-level segmentation.}
    \label{fig:resnetggca_unet}
\end{figure*}

Figure \ref{fig:resnetggca_unet} illustrates the overall architecture of the proposed GCA-ResUNet network. The network is based on the classical U-Net framework and consists of four main components: an encoder, a bottleneck module, a decoder, and skip connections, exhibiting a typical symmetrical encoder–decoder design. In the encoder, we adopt ResNet50 as the backbone feature extractor, replacing the shallow stacked convolutional blocks used in the traditional U-Net. The residual structures in the encoder effectively alleviate the gradient vanishing problem in deep networks while enhancing the model’s capacity to capture complex and fine-grained features in medical images.

Specifically, the encoder first employs a large convolution kernel (7×7) to perform initial feature extraction, followed by batch normalization (BatchNorm) and ReLU activation for nonlinear mapping. A subsequent max-pooling operation reduces the spatial resolution and increases translation invariance. The encoder then passes through four sequential residual layers, generating multi-scale feature maps (feat1–feat5). These feature maps preserve both rich semantic information and spatial details, providing multi-scale support for the subsequent decoder. Each residual layer performs progressive downsampling, enabling the network to capture global semantic context in deeper layers while retaining local details in shallower layers, which is crucial for high-precision medical image segmentation.

The decoder follows a U-Net style upsampling strategy using unetUp modules to progressively restore spatial resolution. Each upsampling unit first applies bilinear interpolation to the low-resolution feature maps, which are then concatenated along the channel dimension with the corresponding encoder feature maps. The concatenated features are further processed through two convolutional layers with ReLU activation, effectively fusing semantic and spatial information. For the ResNet50 backbone, an additional convolutional layer is applied at the final stage of the decoder to further enhance the fused features, ensuring effective integration of high-level semantic information and low-level spatial details. Finally, a 1×1 convolution maps the fused features to the target classes, achieving end-to-end pixel-wise segmentation.

The overall design balances deep semantic representation and spatial localization accuracy, making it particularly suitable for medical images with blurred boundaries, complex object structures, or varying scales. The use of skip connections ensures that shallow features are fully utilized during feature reconstruction, effectively mitigating spatial information loss and improving the fine-grained consistency of segmentation outputs. This encoder–decoder symmetrical design enables GCA-ResUNet to achieve high-precision pixel-level predictions in medical image segmentation tasks, providing a solid foundation for the subsequent detailed description of the ResNet50 backbone module.

\subsection{Backbone}

 The residual network (ResNet) is a landmark in deep learning, as it effectively mitigates the vanishing gradient problem in deep neural networks through residual connections. The basic unit of ResNet50 can be expressed as:
\begin{equation}
y = F(x, \{W_i\}) + W_s x.
\end{equation}
where \(\mathbf{x}\) and \(\mathbf{y}\) denote the input and output of a residual block, respectively, and \(\mathcal{F}(\cdot)\) represents the residual mapping, typically implemented with a sequence of convolutions and batch normalization layers.

We adopt ResNet50 as the base network and modify it as the primary feature extraction backbone. The original ResNet50 is a deep residual network designed for image classification, producing a single feature vector via global average pooling for final class prediction, which demonstrates strong performance on large-scale classification datasets such as ImageNet. However, for tasks requiring multi-scale features and global semantic modeling—such as object detection, semantic segmentation, OCR, and medical image analysis—the single-scale global features are often insufficient. Shallow layers capture rich low-level details but lack high-level semantics, while deep layers provide strong semantic information but lose fine-grained details. Without effectively leveraging multi-level features, models encounter performance bottlenecks in detecting small objects, parsing complex scenes, or processing high-resolution images.

To overcome these limitations, we systematically modify ResNet50. As illustrated in Figure \ref{fig:resnetggca_unet}, the network does not merely output the final classification vector; instead, it retains five intermediate feature maps \([ \text{feat}_1 - \text{feat}_5 ]\), capturing a multi-scale hierarchical representation.
\begin{itemize}
    \item \(feat1\): after the initial convolution and activation,
    \item \(feat2\): output of the first stage,
    \item \(feat3\): output of the second stage,
    \item \(feat4\): output of the third stage,
    \item \(feat5\): output of the fourth stage.
\end{itemize}

This design enables seamless integration with hierarchical feature structures, such as feature pyramid networks (FPN)\cite{lin2017fpn} or U-Net. In addition, a Grouped Coordinate Attention (GCA) module (see \ref{GCA}) is incorporated into each residual block (Bottleneck). The GCA module is applied after the third \(1\times1\) convolution and batch normalization, but before addition with the residual branch. This placement allows GCA to modulate global context on high-dimensional, semantically rich features, enhancing feature representation immediately after local convolution and enabling long-range dependency modeling beyond the convolutional receptive field. This design ensures a balance between local feature extraction and global modeling, improving both discriminative capability and generalization.

The modified backbone preserves the original ResNet50 structure before the residual layers, including the initial \(7\times7\) convolution, max-pooling layer, and four residual layers (Layer1–Layer4) composed of Bottleneck blocks. For downstream tasks, the original global average pooling layer (self.avgpool) and fully connected layer (self.fc) are removed by \texttt{del model.avgpool} and \texttt{del model.fc}. This modification ensures the network outputs spatially-preserved feature maps instead of a single classification vector, which facilitates multi-task prediction.

These structural changes maintain compatibility with the original parameter organization, allowing pre-trained ImageNet\cite{deng2009imagenet} weights to be loaded with strict=False. Matching convolutional and BN layer parameters are retained, while the newly introduced GCA module is randomly initialized and the deleted fully connected layer is ignored. This strategy ensures effective initialization, fast convergence, and adaptability of the new module.
At the residual block level, the classic Bottleneck structure—“\(1\times1\) convolution for dimensionality reduction \(\rightarrow 3\times3\) convolution for local feature extraction \(\rightarrow 1\times1\) convolution for dimensionality expansion”—is retained, with the GCA module inserted at the final stage. As shown in the forward pass, features after convolution and BN are first modulated by GCA and then added to the residual branch. This design provides two key advantages: (1) local convolution and global attention features are fused before residual addition, mitigating limited receptive field issues; (2) the residual mechanism is preserved, retaining ResNet’s training stability and gradient propagation benefits. Consequently, the network extracts edge and texture information in shallow layers, captures semantic categories and contextual relationships in deep layers, and enhances features with global context at each residual stage, substantially improving feature robustness.

Finally, the network outputs a set of multi-level feature maps \([feat1 - feat5]\) that retain rich spatial information, suitable for downstream tasks such as object detection and semantic segmentation. Compared with the original ResNet50, the modified network exhibits clear advantages in output format, global modeling, terminal design, and pre-training compatibility, inheriting ResNet50’s efficiency and stability while addressing its limitations in multi-scale representation and global information modeling.

\subsection{Grouped Coordinate Attention}\label{GCA}

\begin{figure*}[t]
    \centering
    \includegraphics[width=1\textwidth,height=0.9\textheight,keepaspectratio,trim=0 0 0 0,clip]{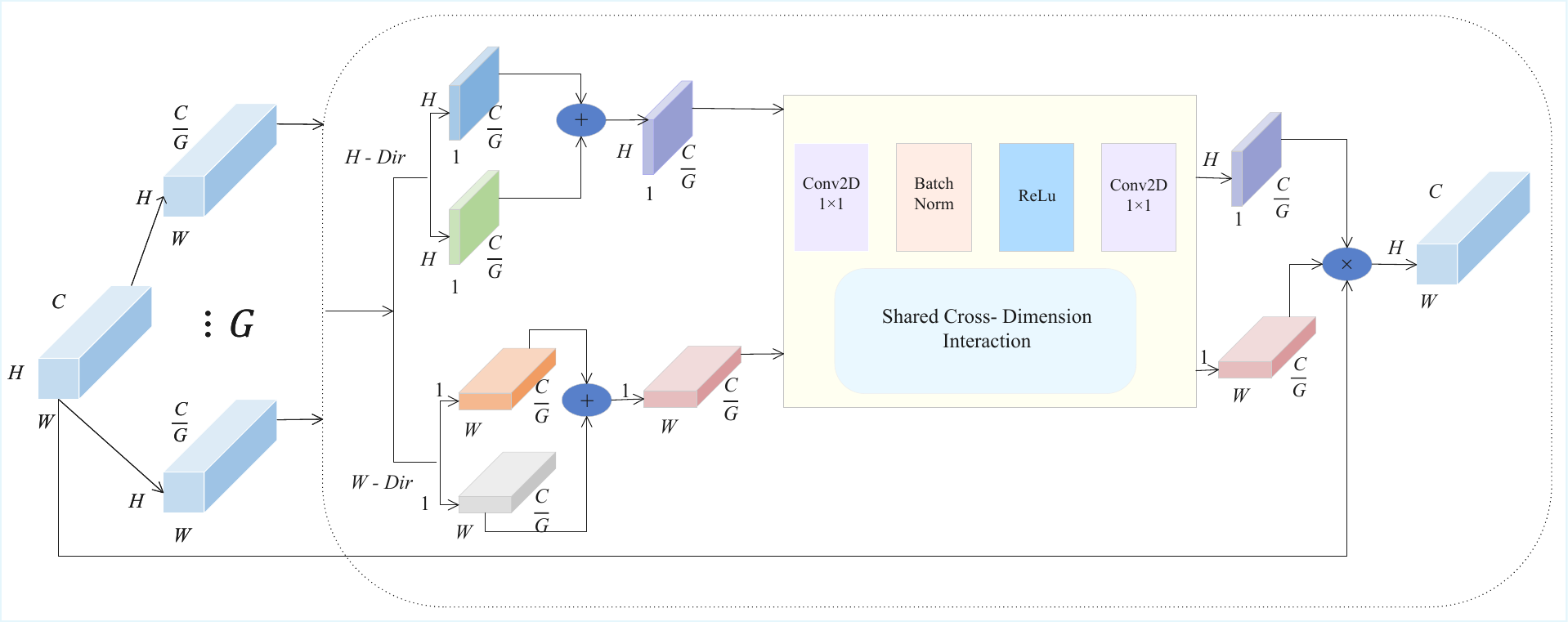}
    \caption{Grouped Coordinate Attention (GCA) network model diagram}
    \label{fig:GGCA}
\end{figure*}

We propose a Grouped Coordinate Attention (GCA) mechanism, as illustrated in Figure \ref{fig:GGCA}, which simultaneously models inter-channel dependencies and directional spatial distribution of features, thereby enhancing the network's ability to capture critical information. Unlike conventional channel attention mechanisms (e.g., SE module) that rely solely on global average pooling to compress the spatial dimension, GCA performs global average pooling and max pooling along both horizontal and vertical directions, effectively preserving positional information and boundary features.

Let the input feature be \(X \in \mathbb{R}^{B \times C \times H \times W}\), where \(B\) denotes the batch size, \(C\) the number of channels, and \(H\) and \(W\) the height and width of the feature map, respectively. We first divide the input channels into \(G\) groups, each of size \(C_g = C / G\). This grouping strategy reduces computational cost and mitigates redundancy across channels. For each group feature \(X_g\in \mathbb{R}^{B \times C_g \times H \times W}\), we perform horizontal pooling by keeping the height \(H\) unchanged and compressing the width to 1:
The grouped feature map $X_g$ undergoes both horizontal and vertical directional pooling operations to capture long-range dependencies along different spatial axes.

\begin{equation}
f^{h}_{avg} = \text{AvgPool}_h(X_g), \quad 
f^{h}_{avg} \in \mathbb{R}^{B \times C_g \times H \times 1}.
\end{equation}
In this step, $f^{h}_{avg}$ aggregates contextual responses across the width dimension through average pooling, providing a smooth estimation of horizontal global features.

\begin{equation}
f^{h}_{max} = \text{MaxPool}_h(X_g), \quad 
f^{h}_{max} \in \mathbb{R}^{B \times C_g \times H \times 1}.
\end{equation}
By contrast, $f^{h}_{max}$ focuses on salient activations along the same axis, enabling the model to emphasize strong horizontal responses and boundary details.

\begin{equation}
f^{w}_{avg} = \text{AvgPool}_w(X_g), \quad 
f^{w}_{avg} \in \mathbb{R}^{B \times C_g \times 1 \times W}.
\end{equation}
To encode vertical context, average pooling is performed along the height dimension, yielding $f^{w}_{avg}$ that captures the overall semantic tendency across rows.

\begin{equation}
f^{w}_{max} = \text{MaxPool}_w(X_g), \quad 
f^{w}_{max} \in \mathbb{R}^{B \times C_g \times 1 \times W}.
\end{equation}
Likewise, $f^{w}_{max}$ extracts the most responsive vertical structures, supplementing the average-pooled features with strong activation cues.
The fused features \(f^h\) and \(f^w\) are then fed into a shared bottleneck convolution network composed of two \(1 \times 1\) convolutions, batch normalization (BN) layers, and ReLU activation. The first convolution reduces the channel dimension to \(C_g / r\) and the second restores it to \(C_g\), where \(r\) is the channel reduction ratio:
\begin{align}
Z &= \sigma\Big(\text{Conv}_{1 \times 1}(\delta(BN(\text{Conv}_{1 \times 1}(F))))\Big)
\end{align}Here, \(\delta(\cdot)\) denotes the ReLU activation function, and \(\sigma(\cdot)\) denotes the Sigmoid activation function. 
Through this process, we obtain the attention weights along the horizontal and vertical directions.
\begin{equation}
A^{h} \in \mathbb{R}^{B \times C_g \times H \times 1},
\end{equation}
where \(A^{h}\) represents the attention map along the horizontal direction, capturing contextual dependencies across the height dimension.
\begin{equation}
A^{w} \in \mathbb{R}^{B \times C_g \times 1 \times W},
\end{equation}
where \(A^{w}\) denotes the attention map along the vertical direction, modeling long-range relationships across the width dimension.
Finally, the directional attention weights are applied to the original input feature via broadcasting, enabling the joint modeling of global and local dependencies. Specifically, the output of the GCA module can be formulated as:
\begin{equation}
Y_g = X_g \otimes A^{h} \otimes A^{w},
\end{equation}
where \(\otimes\) denotes element-wise multiplication. The final output feature is obtained by concatenating the grouped results:
\begin{equation}
Y = \text{Concat}(Y_1, Y_2, \ldots, Y_G).
\end{equation}
Here, $\odot$ denotes element-wise multiplication, $Y_g$ represents the weighted feature of the $g$-th group, and the final output $Y$ is obtained by concatenating all groups. This mechanism not only effectively captures long-range dependencies but also maintains sensitivity to spatial directions with minimal computational overhead. Compared with the SE module, GCA no longer compresses features into a one-dimensional channel descriptor; instead, it preserves significant spatial information through decomposed modeling. Compared with CBAM, GCA avoids the additional convolution overhead for spatial attention, achieving more efficient modeling via directional pooling and shared convolutions. Compared with CoordAttention, GCA introduces a grouping strategy while retaining coordinate information, providing stronger discriminative capability and robustness at low computational complexity. Overall, GCA adds only a minimal number of parameters while significantly enhancing feature representation, providing richer support for downstream representation learning tasks.

Integrating GCA into the ResNet50 Bottleneck structure significantly improves feature representation. In a standard residual block, features are directly propagated after convolutional extraction. The introduction of GCA provides direction-aware global attention modulation to the Bottleneck without significantly increasing the number of parameters. Its advantages are twofold: first, it enhances the network’s ability to jointly model local details and global structure, making residual blocks more sensitive to complex textures or boundary regions; second, the grouping strategy controls computational complexity, allowing the module to remain efficient in large-scale networks. Experimental results demonstrate that embedding GCA into ResNet50 can effectively improve the accuracy of downstream segmentation tasks while maintaining model lightweightness, particularly in preserving boundary details and recognizing small target regions.

\subsection{Decoder}

In the decoder, to progressively recover the spatial resolution lost during the encoder’s downsampling and to integrate multi-level feature information, we employ an upsampling module composed of bilinear interpolation, feature concatenation, and a two-layer convolution--activation block.

Specifically, the low-resolution feature input $\text{inputs}_2$ is first spatially upsampled using \texttt{UpsamplingBilinear2d}, doubling its dimensions to match the high-resolution feature $\text{inputs}_1$ from the corresponding encoder layer. Subsequently, the two feature maps are concatenated along the channel dimension, effectively fusing high-level semantic information with low-level fine-grained details.

The concatenated feature map is then passed through two successive $3 \times 3$ convolution layers with ReLU activations. The first convolution maps the high-dimensional concatenated features to the target number of channels, achieving channel reduction and preliminary feature fusion. The second convolution preserves the channel dimension while further enhancing the local semantic representation and spatial details of the fused features, thereby improving the decoder’s capacity for boundary delineation and small object segmentation.

Compared to conventional transposed convolutions, the bilinear interpolation employed in this module effectively mitigates checkerboard artifacts, generating smoother feature maps with lower computational cost. This lightweight design allows the network to maintain both segmentation accuracy and inference efficiency when processing high-resolution medical images. Coupled with attention-enhanced feature representations in the encoder (e.g., the GCA module), the decoder can achieve high-precision segmentation while controlling the number of parameters and computational overhead.

\section{Experiment}
\subsection{Datasets}
\textbf{Synapse Multi-Organ Segmentation Dataset (Synapse).} The Sy-
napse dataset is a widely used public benchmark for abdominal multi-organ segmentation. It contains 30 abdominal clinical CT scans, comprising a total of 3,779 axial slices. All images are manually annotated at the pixel level by expert radiologists, covering eight major abdominal organs: the aorta, gallbladder, spleen, left kidney, right kidney, liver, pancreas, and stomach. This dataset is extensively used to evaluate the performance of multi-organ automatic segmentation methods and has significant research and clinical value. In this study, we employ the average Dice Similarity Coefficient (DSC) as the primary evaluation metric to quantify the segmentation accuracy for each organ. Due to challenges such as low tissue contrast, large anatomical variations, and unclear organ boundaries in abdominal CT images, the Synapse dataset poses considerable difficulty for segmentation algorithms, effectively testing model robustness in complex medical imaging scenarios. To mitigate bias caused by the limited number of samples, we adopt a multi-fold cross-validation strategy, which reduces the dependency of model performance on a single data split, significantly decreases the variance of the evaluation results, and ensures the stability and reliability of the experiments.

\textbf{Automated Cardiac Diagnosis Challenge Dataset (ACDC).} The ACDC dataset contains cardiac MRI scans from 100 patients, with images annotated by experts to include three major structures: the left ventricle (LV), right ventricle (RV), and myocardium (Myo). Following existing study protocols, the dataset is divided into three subsets for training, validation, and testing. The Dice Similarity Coefficient (DSC) is used as the primary metric to evaluate model performance.
\begin{table}[t]
    \centering
    \caption{Performance comparison on the Synapse multi-organ dataset.}
    \label{tab:synapse_results}
    \resizebox{\textwidth}{!}{ 
    \begin{tabular}{l *{9}{c}}
    \hline
    \textbf{Methods} & \textbf{DSC} & \textbf{Aorta} & \textbf{Gallbladder} & \textbf{Kidney(L)} & \textbf{Kidney(R)} & \textbf{Liver} & \textbf{Pancreas} & \textbf{Spleen} & \textbf{Stomach} \\
    \hline
    R50 U-Net\cite{chen2024transunet} & 77.61 & 87.74 & 63.66 & 80.60 & 78.19 & 93.74 & 56.90 & 85.87 & 74.16 \\
    R50 Att-UNet\cite{chen2024transunet} & 75.57 & 55.92 & 63.91 & 79.20 & 72.71 & 93.56 & 49.37 & 87.19 & 74.95 \\
    U-Net\cite{ronneberger2015unet} & 76.85 & 89.07 & 69.72 & 77.77 & 68.60 & 93.43 & 53.98 & 86.67 & 75.58 \\
    Att-U-Net\cite{oktay2018attentionunet} & 77.77 & 89.55 & 68.88 & 77.98 & 71.11 & 93.57 & 58.04 & 87.30 & 75.75 \\
    MT-UNet\cite{wang2022mixedunet} & 78.59 & 87.92 & 64.99 & 81.47 & 77.29 & 93.06 & 59.46 & 87.75 & 76.81 \\
    SwinUnet\cite{cao2022swinunet} & 79.13 & 85.47 & 66.53 & 83.28 & 79.61 & 94.29 & 56.58 & 90.66 & 76.60 \\
    SelfReg-UNet\cite{zhu2024selfregunet} & 80.54 & 86.07 & 69.65 & 85.12 & 82.58 & 94.18 & 61.08 & 87.42 & 78.22 \\
    VM-UNet\cite{ruan2024vmunet} & 81.08 & 86.40 & 69.41 & 86.16 & 82.76 & 94.17 & 58.80 & 89.51 & 81.40 \\
    \hline
    GCA-ResUNet & \textbf{86.11} & \textbf{89.95} & \textbf{74.71} & \textbf{93.38} & \textbf{91.60} & \textbf{95.54} & \textbf{68.95} & \textbf{92.60} & \textbf{82.15} \\
    \hline
    \end{tabular}
    }
\end{table}
\begin{figure}[t]
    \centering
    \includegraphics[width=\linewidth]{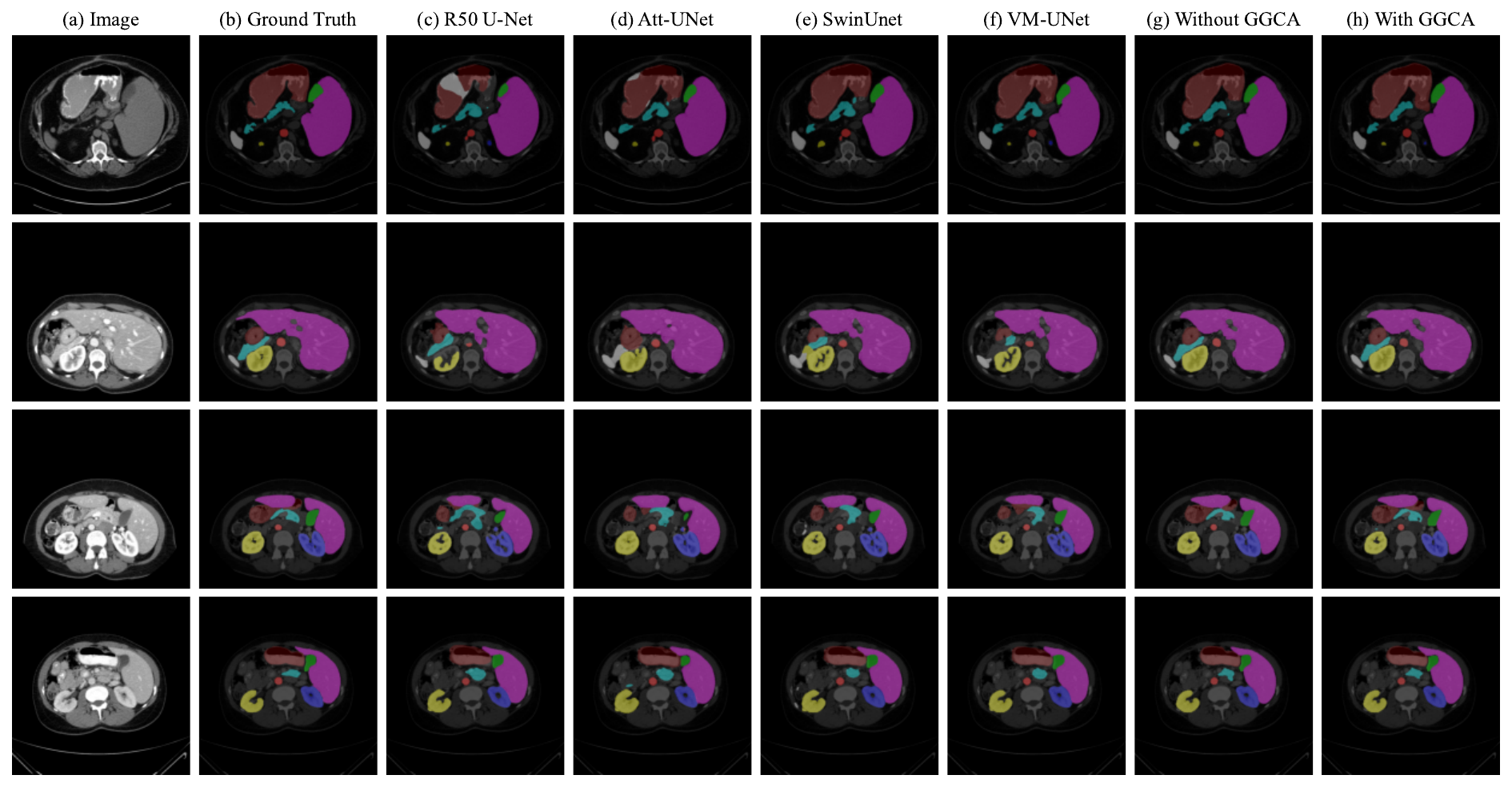} 
    \caption{Comparison of segmentation performance on the Synapse dataset}
    \label{fig:output1}
\end{figure}
\begin{table}[!t]
    \centering
    \caption{Performance comparison on the ACDC dataset.}
    \label{tab:ACDC_results}

    \setlength{\tabcolsep}{1pt} 
    \small 

    \begin{tabular}{l c c c c}
        \hline           
        \textbf{Methods} & \textbf{DSC} & \textbf{RV} & \textbf{Myo} & \textbf{LV} \\
        \hline
        R50 Att-UNet\cite{chen2024transunet}   & 86.75 & 87.58 & 79.20 & 93.47 \\
        R50 U-Net\cite{chen2024transunet}      & 87.55 & 87.10 & 80.63 & 94.92 \\
        U-Net\cite{ronneberger2015unet}        & 89.68 & 87.17 & 87.21 & 94.68 \\
        SwinUnet\cite{cao2022swinunet}        & 90.00 & 88.55 & 85.62 & 95.83 \\
        MT-UNet\cite{wang2022mixedunet}        & 90.43 & 86.64 & 89.04 & 95.62 \\
        SelfReg-UNet\cite{zhu2024selfregunet}  & 91.49 & 89.49 & 89.27 & 95.70 \\
        \hline
        GCA-ResUNet                            & \textbf{92.64} & \textbf{92.27} & \textbf{89.37} & \textbf{96.30} \\
        \hline
    \end{tabular}
\end{table}
\begin{figure}[!t]
    \centering
    \includegraphics[width=\linewidth]{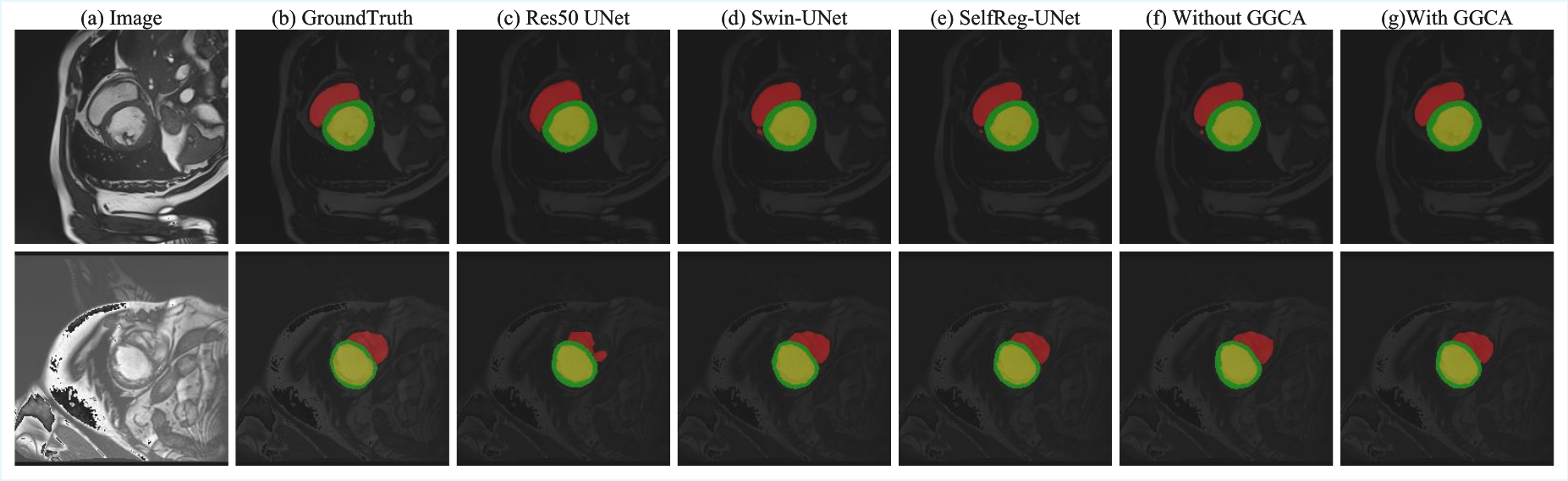} 
    \caption{Comparison of segmentation performance on the ACDC dataset}
    \label{fig:output2}
\end{figure}

\subsection{Ablation Study}
\begin{figure}[!t]
    \centering
    \includegraphics[width=\linewidth]{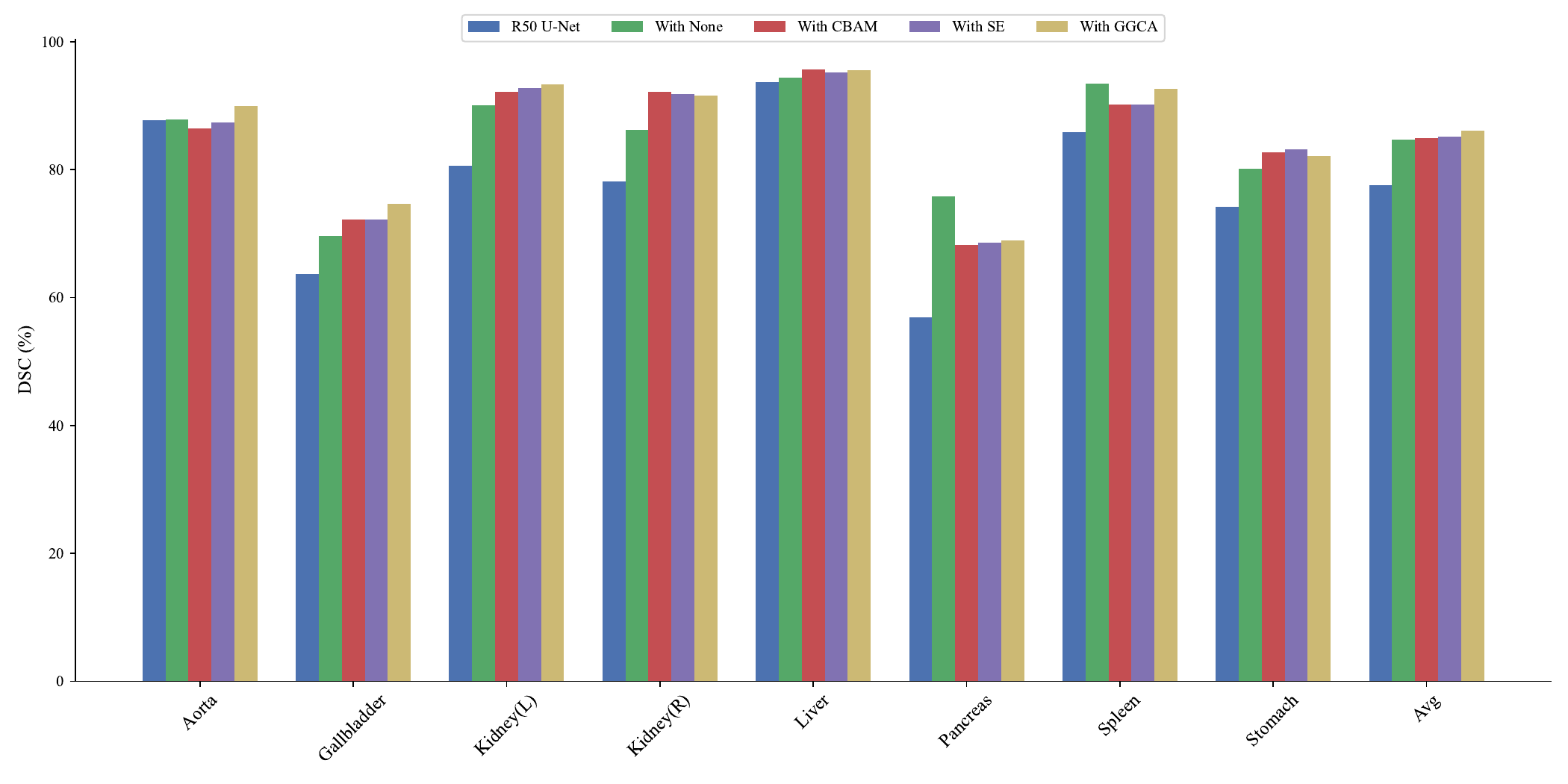} 
    \caption{Performance Comparison of Original and Modified ResNet-UNet with Different Modules on Synapse Dataset}
    \label{fig:Synapse_Ablation}
\end{figure}

\begin{figure}[!t]
    \centering
    \includegraphics[width=0.6\linewidth]{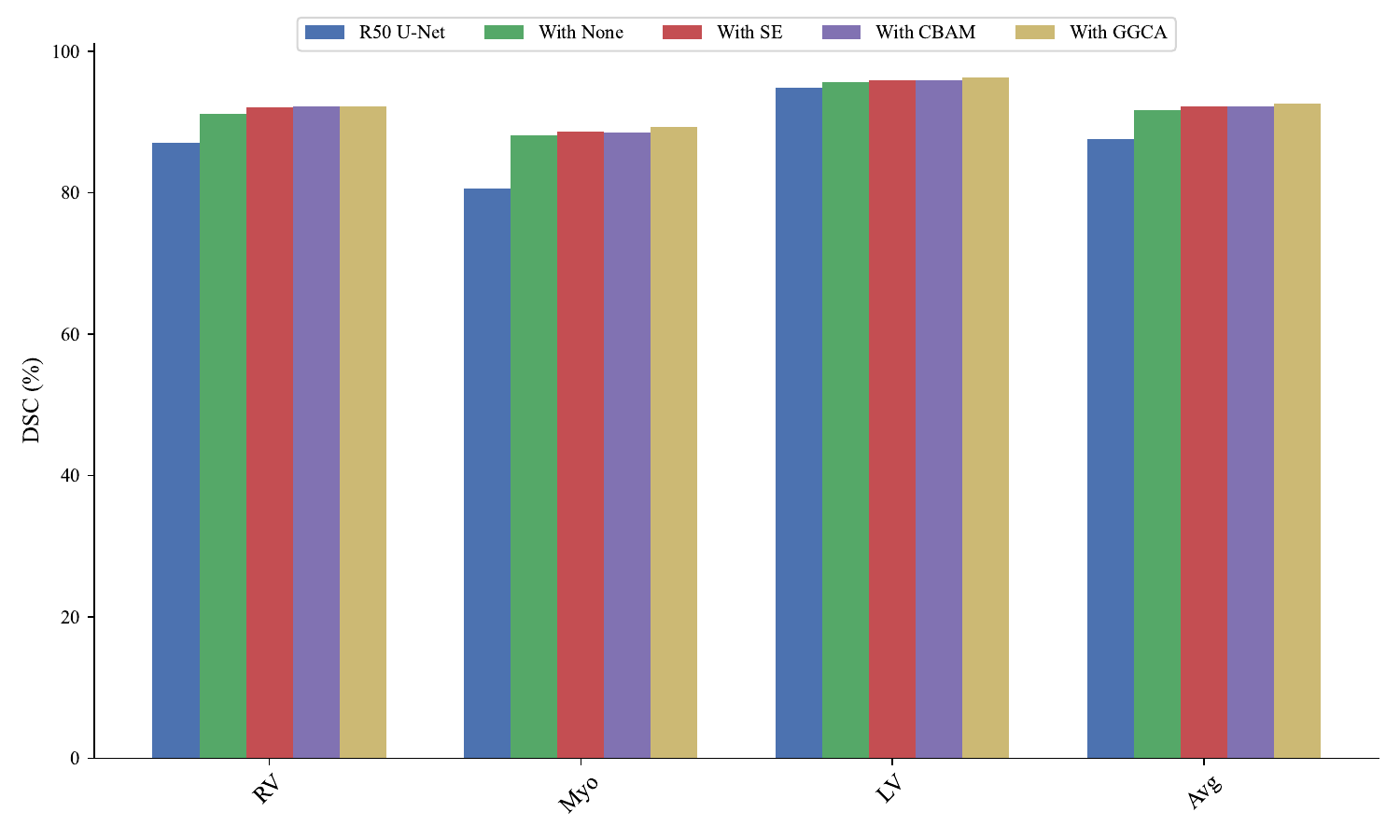} 
    \caption{Performance Comparison of Original and Modified ResNet-UNet with Different Modules on ACDC Dataset}
    \label{fig:ACDC_Ablation}
\end{figure}

To systematically evaluate the effectiveness of the proposed GCA module in medical image segmentation tasks, a series of ablation experiments were conducted on the modified ResNet50-UNet backbone. The experiments first compared the original ResNet50-UNet with the modified ResNet50, and further examined the performance differences when the GCA module was incorporated, thereby assessing its contribution to enhancing global feature representation. Additionally, to analyze the effects of different attention mechanisms within the residual network, SE (Squeeze-and-Excitation), CBAM (Convolutional Block Attention Module), and the GCA modules were individually integrated for comparison under identical network structures and training conditions. All experiments employed the Dice Similarity Coefficient (DSC) as the primary evaluation metric to quantify the contribution of each module to segmentation performance.Specifically illustrated in Figure \ref{fig:Synapse_Ablation} and Figure \ref{fig:ACDC_Ablation}.

\subsection{Training}
The proposed network was evaluated on two datasets to assess its segmentation performance. All experiments were conducted under a consistent training setup: input images were uniformly resized to 224×224(the same size as those in \cite{chen2024transunet,ruan2024vmunet,zhu2024selfregunet}, and so on), with a batch size of 8, and the Adam optimizer was employed with an initial learning rate of 1$\times$10$^{-4}$. The total loss function was defined as a combination of Dice Loss and Cross-Entropy (CE) Loss. Dice Loss measures the overlap between the predicted segmentation and ground-truth labels, enhancing robustness to class imbalance and ambiguous boundaries, while CE Loss ensures pixel-wise classification accuracy. The combination of these losses enables the model to simultaneously optimize overall segmentation quality and pixel-level precision. The network was trained from scratch without using pretrained ResNet50 weights to verify its feature learning capability and generalization performance. All experiments were performed on a single NVIDIA RTX 4060 Ti GPU (16 GB memory), with actual memory consumption below 4 GB. Under the same experimental conditions, the proposed approach was compared with several representative U-Net variants to comprehensively evaluate its segmentation accuracy and computational efficiency.

\section{Conclusion}
In this study, we systematically improved the classical ResNet50 network for semantic segmentation tasks. By adjusting the downsampling strategy and interface design, the network was efficiently integrated with U-Net’s skip connections, and the Grouped Coordinate Attention (GCA) was incorporated into the Bottleneck modules to enhance the network’s ability to model long-range dependencies and global context. Based on this design, the modified backbone effectively captures both low-level details and high-level semantic information, significantly improving boundary delineation and small-object recognition.

Experimental results demonstrate that the proposed method outperforms the conventional ResNet50 and typical U-Net variants across multiple public datasets, validating its advantages in feature representation richness, contextual understanding, and adaptability to downstream tasks. Furthermore, the network maintains high segmentation accuracy while keeping computational costs low.

In summary, this study presents a lightweight and efficient ResNet50 enhancement, providing an improved backbone choice for semantic segmentation tasks. Future work will further explore the application potential of GCA in multi-task learning and 3D medical image segmentation, combined with lightweight design and self-supervised pretraining strategies, to further enhance model generalization and applicability.

\bibliographystyle{elsarticle-num-names}
\bibliography{references}
\end{document}